\documentclass[runningheads]{llncs}
\usepackage{graphicx}

\usepackage{tikz}
\usepackage{comment}
\usepackage{amsmath,amssymb} 
\usepackage{color}

\usepackage{url}
\usepackage{hyperref}
\definecolor{RepoBlue}{rgb}{0.23,0.49,0.72}
\definecolor{ReferenceBlue}{rgb}{0.07,0.18,0.43}
\definecolor{FigureTableRed}{rgb}{0.53,0.12,0.17}
\hypersetup{colorlinks=true,linkcolor=black,citecolor=ReferenceBlue,urlcolor=black}
\newcommand*{\figtabref}[1]{\hyperref[#1]{\textcolor{FigureTableRed}{\ref*{#1}}}}

\begin{document}

\pagestyle{headings}
\mainmatter

\title{Learning Delicate Local Representations for Multi-Person Pose Estimation} 

\titlerunning{Learning Delicate Local Representations for Multi-Person Pose Estimation} 
\authorrunning{Yuanhao Cai$^{*}$, Zhicheng Wang$^{*}$ et al.} 

\author{
    Yuanhao Cai$^{1,2}$\thanks{The first two authors contribute equally to this work. This work is done when Yuanhao Cai, Zhengxiong Luo, Binyi Yin and Angang Du  are interns at Megvii Research.} \and Zhicheng Wang$^{1*}$ \and Zhengxiong Luo$^{1,3}$  \and Binyi Yin$^{1,4}$\and \\Angang Du$^{1,5}$  \and Haoqian Wang$^{2}$ \and Xiangyu Zhang$^{1}$ \and \\Xinyu Zhou$^{1}$\and Erjin Zhou$^{1}$  \and Jian Sun$^{1}$ \\
}
\institute{$^1$Megvii Inc. \quad 
    $^2$Tsinghua University \quad 
    $^3$Chinese Academy of Sciences \quad \\
    $^4$Beihang University \quad
    $^5$Ocean University of China \\
    $^{1}$\textit{\{caiyuanhao, wangzhicheng, zxy, zej, zhangxiangyu, sunjian\}@megvii.com}\\
    $^{2}$\textit{wanghaoqian@tsinghua.edu.cn}
    }
    
\maketitle

\begin{abstract}
In this paper, we propose a novel method called Residual Steps Network~(RSN). RSN aggregates features  with the same spatial size~(Intra-level features) efficiently to obtain delicate local representations, which retain rich low-level spatial information and result in precise keypoint localization. Additionally, we observe the output features contribute differently to final performance. To tackle this problem, we propose an efficient attention mechanism - Pose Refine Machine~(PRM) to make a trade-off between local and global representations in output features and further refine the keypoint locations. Our approach won the 1st place of COCO Keypoint Challenge 2019 and achieves state-of-the-art results on both COCO and MPII benchmarks, without using extra training data and pretrained model. Our single model achieves 78.6 on COCO test-dev, 93.0 on MPII test dataset. Ensembled models achieve 79.2 on COCO test-dev, 77.1 on COCO test-challenge dataset. The source code is publicly available for further research at {\hypersetup{urlcolor=RepoBlue}\url{https://github.com/caiyuanhao1998/RSN/}}

\end{abstract}

\section{Introduction}
The goal of multi-person pose estimation is to locate keypoints of all persons in a single image. It is a fundamental task for human motion recognition, kinematics analysis, human-computer interaction, animation etc. For years, human pose estimation was based on handcraft features. Recently, It has made great progress with the development of deep convolutional neural network. 
\begin{figure}[h]
 \begin{center}
 \includegraphics[width=1.0\textwidth]{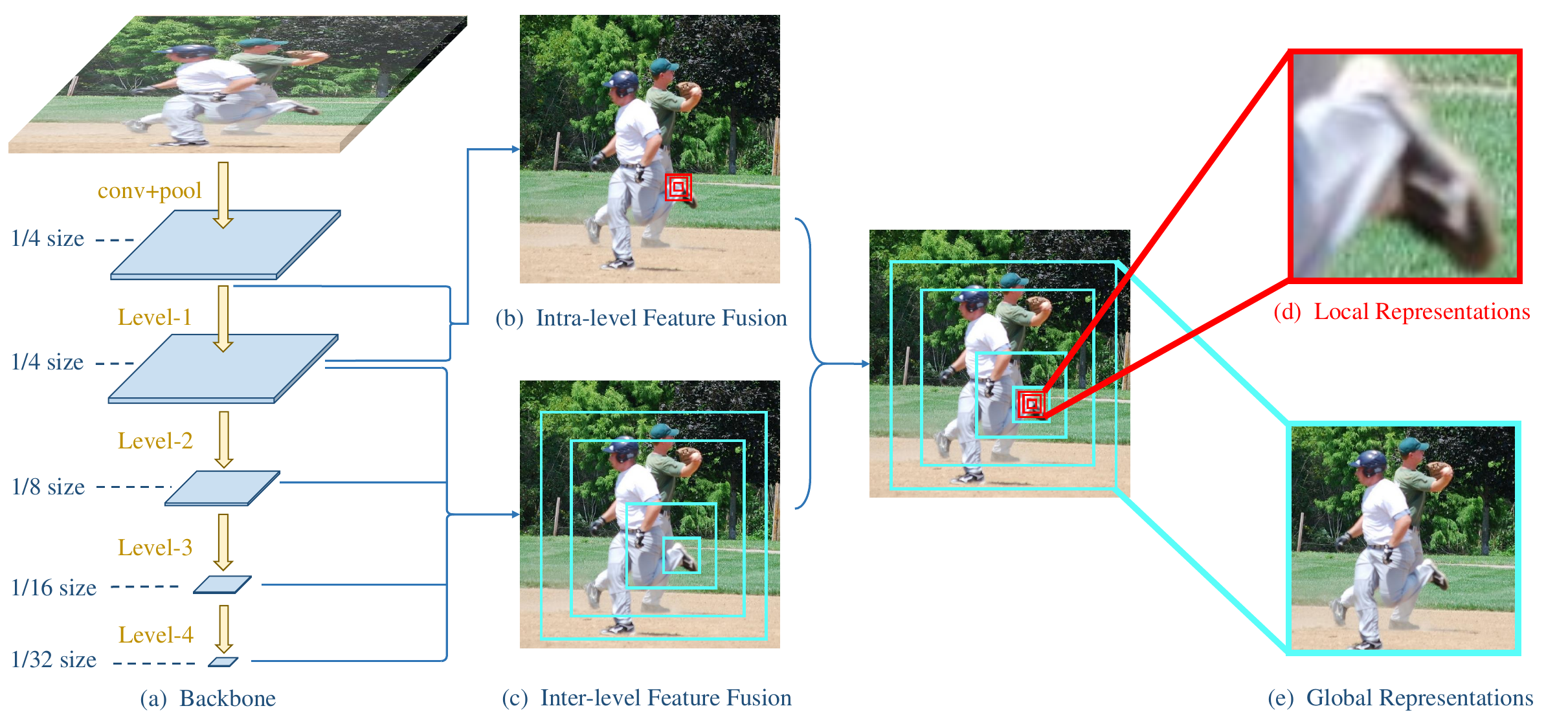} 
 \caption{Comparison of intra-level feature fusion and inter-level feature fusion. (a) Backbone. "1/4 size" means 1/4 size of input image. (b) Intra-level feature fusion of level 1. (c) Inter-level feature fusion. (d) Local Representations. (e) Global representations.} 
 \label{fig:human} 
 \end{center}
\end{figure}
The task of human pose estimation concerns both keypoint localization and classification. Spatial information benefits the localization task, while semantic information is good for the classification task. To extract these two kinds of information, current methods mainly focus on aggregating inter-level features. For instance, HRNet~\cite{hrnet} maintains spatial information in high-resolution sub-network and gradually adds semantic information to it from low-resolution sub-networks. In this way, features of different levels are fully aggregated. In CPN~\cite{cpn}, features of four different spatial levels are extracted by the backbone, and they are combined by a head network. Although these methods are different in the ways of feature fusion, the features to be aggregated are always from different levels. On the contrast, the feature fusion within the same level stays less explored in the task of human pose estimation. 

The comparison of intra-level feature fusion (level 1) and inter-level feature fusion is illustrated in Figure~\figtabref{fig:human}. The feature maps are continuously downsampled to 1/4, 1/8, 1/16, 1/32 size of input image in Figure~\figtabref{fig:human}(a). We define consecutive feature maps with the same spatial size as one level. As Figure~\figtabref{fig:human}(c) depicts, there is a big gap between the receptive fields of features from different levels, which are indicated by light blue bounding boxes. As a result, representations learned by inter-level feature fusion are relatively coarse, which impede the localization of human pose from precise. As Figure~\figtabref{fig:human}(b) shows, the gap between the receptive fields of intra-level features which are indicated by red bounding boxes is relatively small. As shown in Figure~\figtabref{fig:human}(d), fusing intra-level features can extract much more delicate local representations retaining more precise spatial information, which is critical to keypoint localization.

To learn better local representations, we propose a novel network architecture - Residual Steps Network~(RSN). The Residual Steps Block~(RSB) of RSN fuses features inside each level using dense element-wise sum operations, which is shown in Figure~\figtabref{fig:pipeline}(c).
The inner structure of RSB is deeply connected and motivated by DenseNet~\cite{densenet}, which has a good performance for human pose estimation owing to retaining rich low-level features by deep connections. However, deep connections bring about explosion of the network capacity as it goes deeper. Thus, DenseNet performs poorly when the network becomes large. RSN is motivated by DenseNet but is quite different in that RSN uses element-wise sum rather than concatenation to circumvent network capacity explosion. RSN is modestly less dense connected in the block than DenseNet, which further promotes the efficiency. Additionally, we observe that the output features containing both global and local representations contribute differently to final performance. In light of this observation, we propose an attention module - Pose Refine Machine (PRM) to rebalance the output features of the network. The architecture of PRM is illustrated in Figure~\figtabref{fig:RB&FB} and analyzed in Section~\ref{sec:RM}. To better illustrate the advantages of our approach, we analyze the differences between RSN and current methods in Section~\ref{sec:fusion}.


In conclusion, our contributions can be summarized as three points:



1. We propose a novel network - RSN, which aims to learn delicate local representations by efficient intra-level feature fusion.

2. We propose an attention mechanism - PRM, which goes further to make a trade-off between local and global representations, and benefits the final performance. 

3. Comprehensive experiments demonstrate that Our approach outperforms the state-of-the-art methods on both COCO and MPII datasets without using extra training data and pretrained model. Moreover, the proposed approach is much faster than HRNet with comparable performance on both GPU and CPU platforms.




\section{Related Work}
\subsection{Multi-person Pose Estimation}
Current methods of human pose estimation fall into two categories: top-down methods~\cite{cpn28,cpn18,mask-rcnn,cpn9,hrnet,hourglass,toutiao,top_1,top_2,cpn,simplebase} and bottom-up methods~\cite{cmu-pose,multipose,bottom_1,bottom2}. Top-down methods first detect the positions of all persons, then estimate the pose of each person. Bottom-up methods first detect all the human keypoints in an image and then assemble these points into groups to form different individuals. 
Since this paper mainly concentrates on feature fusion strategies, we discuss these methods in terms of feature fusion.
\begin{figure*}[h]
 \centering
 \includegraphics[width=1.0\textwidth]{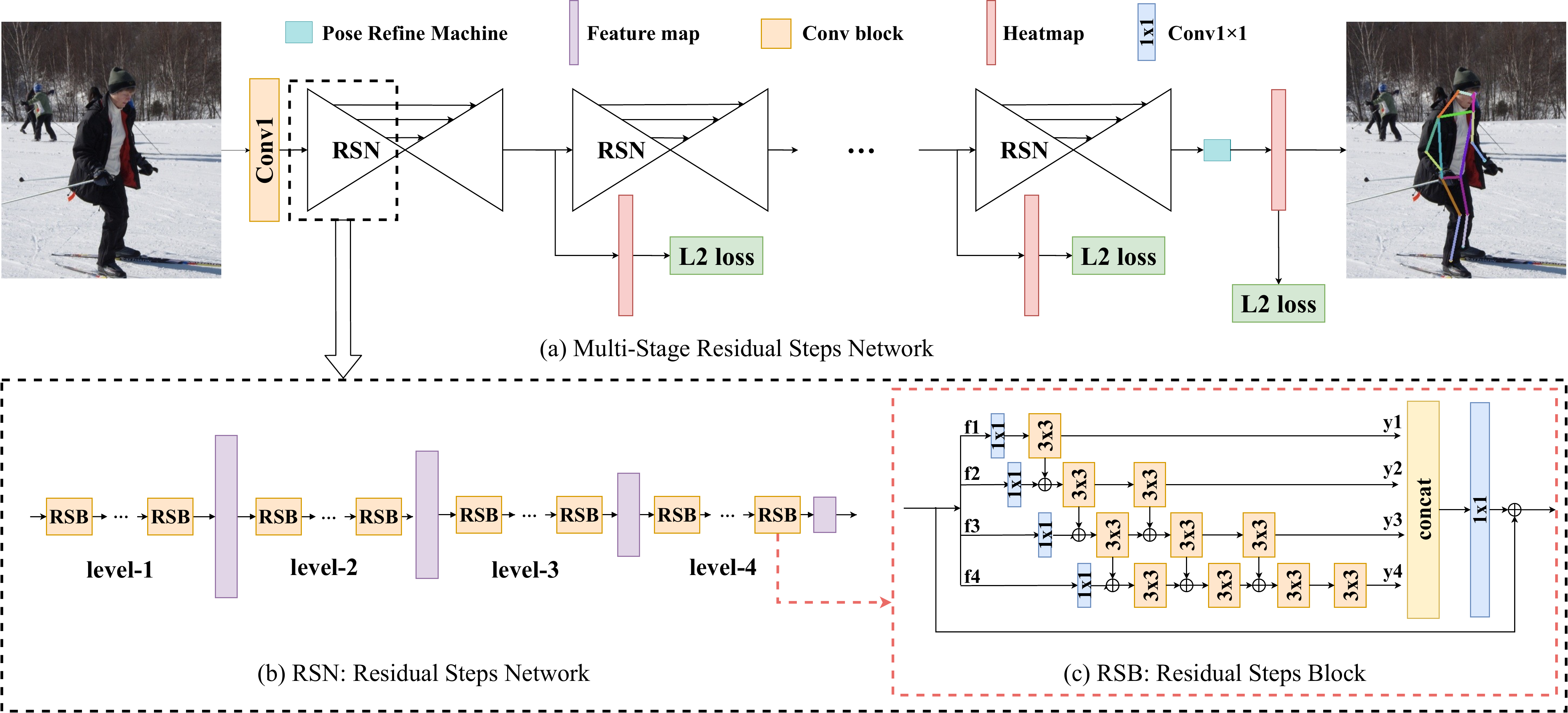} 
 \caption{Our pipeline. (a) is the multi-stage network architecture. It is cascaded by several Residual Steps Networks (RSNs). (b) is the backbone of RSN. (c) is the structure of Residual Steps Block (RSB), which is the basic block of RSN. RSB is designed for learning delicate local representations through dense element-wise sum connections. A Pose Refine Machine (PRM) is used in the last stage and it is analyzed in Section 3.4.
 } 
 \label{fig:pipeline} 
\end{figure*}
\subsection{Feature Fusion}\label{sec:fusion}
Recently, many methods~\cite{hourglass,simplebase,cpn,lpf,hrnet} of human pose estimation use inter-level feature fusion to extract more spatial and semantic information. Newell et al.~\cite{hourglass} propose a U-shape convolutional neural network~(CNN) named Hourglass. In a single stage of hourglass, high-level features are added to low-level features after upsampling. Later works such as Yang et al.~\cite{lpf} show great performance of using inter-level feature fusion. Chen et al.~\cite{cpn} combines inter-level features using a RefineNet. Sun et al.~\cite{hrnet} set up four parallel sub-networks. The features of these four sub-networks aggregate with each other through high-to-low or low-to-high way.

 Though many methods have validates the effectiveness of inter-level feature fusion, intra-level feature fusion is rarely explored in human pose estimation. However, it has extensive applications in other tasks such as semantic segmentation and image classification~\cite{inception,res2net,densenet,dpn,osnet,resnext}. In a block of Inception~\cite{inception}, features pass through several convolutional layers with different kernels separately and then added up.
DenseNet~\cite{densenet} fuses intra-level features using continuous concatenating operations. This implementation retains low-level features to improve the performance. However, when the network goes deeper, the capacity increases sharply and much redundant information appears in the network, resulting in poor efficiency. Different from DenseNet, RSN uses element-wise sum rather than concatenation to circumvent network capacity explosion. In addition, RSN is modestly less densely connected in the constituent unit, which further promotes the efficiency.

Res2Net~\cite{res2net} and OSNet~\cite{osnet} focus on multi-scale representations. Both of them lack dense connections between adjacent branches. The dense connections contribute sufficient gradients and make low-level features better supervised. Therefore, lack of dense connections between adjacent branches results in less precise spatial information, which is essential to keypoint localization. Suffering from this limitations, both Res2Net and OSNet are inferior to RSN in the task of human pose estimation. In Section~\ref{sec:high-efficiency}, we validate the efficiency of DenseNet, Res2Net and RSN. 

\subsection{Attention Mechanism}
Attention mechanism~\cite{CBAM,senet,toutiao,xunit,dfn,pam,pw} is almost used in all areas of computer vision. 
Current methods of attention mechanism mainly fall into two categories: channel attention~\cite{CBAM,senet,toutiao,dfn} and spatial attention~\cite{toutiao,pw,CBAM,xunit,pam}. 
Woo et al.~\cite{CBAM} propose a channel attention module with global average pooling and max pooling. Kligvasser et al.~\cite{xunit} propose a spatial activation function with depth-wise separable convolution. Other works such as Hu et al.~\cite{senet} show the advantages of using attention mechanism. However, most prior attention modules are lack of representing capacity and focus on optimizing the backbone. We design PRM to make a trade-off between local and global representations in output features by using powerful while computation-economical operations.

\section{Proposed Method}
The overall pipeline of our method is illustrated in Figure~\figtabref{fig:pipeline}. The multi-stage network architecture is cascaded by several single-stage modules - Residual Steps Network (RSN), shown in Figure~\figtabref{fig:pipeline}(a). As Figure~\figtabref{fig:pipeline}(b) shows, RSN differs from ResNet in the architecture of constituent unit. RSN consists of Residual Steps Blocks (RSBs) while ResNet is comprised of "bottleneck" blocks. Figure~\figtabref{fig:pipeline}(c) illustrates the structure of RSB. A Pose Refine Machine (PRM) is used in the last stage and it is analyzed in Section~\ref{sec:RM}.


  

\subsection{Delicate Local Representations Learning}
Residual Steps Network is designed for learning delicate local representations by repeatedly enhancing efficient intra-level feature fusion inside RSB, which is the constituent unit of RSN.
As shown in Figure~\figtabref{fig:pipeline}(c), RSB firstly divides the features into four splits~$f_i$~(i = 1, 2, 3, 4), then implements a conv1$\times$1~(convolutional layer with kernel size 1$\times$1) separately. Each feature output from conv1$\times$1 undergoes incremental numbers of conv3$\times$3. The output features~$y_i$ (i = 1, 2, 3, 4) are then concatenated to go through a conv1$\times$1. An identity connection is employed as the ResNet bottleneck. Because the incremental numbers of conv3$\times$3 look like steps, the network is therefore named Residual Steps Network.

The receptive fields of RSB range across several values, and the max one is 15. Compared with a single receptive field in ResNet bottleneck as shown in Table~\figtabref{tb:receptfield}, RSB indicates more delicate information viewed in the network. In addition, it is deeply connected inside RSB. On the $i^{th}$ branch, the front $i-1$ conv3$\times$3 receive the features output from the $(i-1)^{th}$ branch. The $i^{th}$ conv3$\times$3 is then designed to refine the fusion of the features output from the $(i-1)^{th}$ conv3$\times$3. Benefit from the dense connection structure, small-gap receptive fields of features are fully fused resulting in delicate local representations, which retain precise spatial and semantic information.
\begin{table}[h]
  \footnotesize

  \centering
  \caption{The receptive field comparison  between RSB and other methods.}
  \begin{tabular}{c|c|c|c|c}
    \hline\hline
    ~~~~Architecture~~~~ &~~~~~ y1 ~~~~~& ~~~~~y2~~~~~ & ~~~~~~y3~~~~~~ &~~~~~~~~y4~~~~~~~~ \\
    \hline
    ResNet & 3 &3 &3 &3                   \\
    OSNet & 3 &5 &7 &9                   \\
    Res2Net & 1 &3 &3,5 &3,5,7                   \\
    RSN & 3 &5,7 &7,9,11 &9,11,13,15                   \\
    \hline\hline                        
  \end{tabular}
  
  \label{tb:receptfield}
\end{table}
Additionally, during the training process, the deeply connected structure contributes  sufficient gradients, so the low-level features are better supervised, which benefits the keypoint localization task. We investigate how the branch number of RSB influences the prediction results in Section~\ref{sec:branch}. Four-branch architecture has the best performance.

\subsection{Receptive Field Analysis}\label{sec:recept-filed}
In this part, we analyze the receptive fields in RSB and other methods. Firstly, the formula for calculating the receptive field of the $k^{th}$ convolutional layer is written as Equation~\ref{eq2}

\begin{equation}
  l_{k} = l_{k-1}+[(f_{k}-1)*\prod_{i=1}^{k-1}{s_{i}}]
  \label{eq2}
\end{equation}

$l_{k}$ denotes the size of the receptive field corresponding to the $k^{th}$ layer, $ f_{k}$ denotes the kernel size of the $k^{th}$ layer and $ s_{i}$ denotes the stride of the $i^{th}$ layer. In this part, we only focus on the change of relative receptive fields in a block. Every $f_{k}$ is 3 and $s_{i}$ is 1. Thus, Equation~\ref{eq2} can be simplified to Equation~\ref{eq2_simple}

\begin{equation}
  l_{k} = l_{k-1}+2
  \label{eq2_simple}
\end{equation}

Using this formula, we calculate the relative receptive fields of RSB and other methods, as shown in Table~\figtabref{tb:receptfield}. It indicates that RSN has a wider range of scales than ResNet, Res2Net and OSNet.
The scale of each human joint varies a lot. For instance, the scale of eye is small while that of hip is large. For this reason, architecture with wider range of receptive fields is more fit for extracting features relating to different joints. Also, wider range of receptive fields helps to learn more discriminant semantic representations, which benefits the keypoint classification task.
More importantly, RSN builds dense connections between the features with small-gap receptive fields inside RSB. The deeply connected architecture contributes to learning delicate local representations, which are essential to precise human pose estimation.


\begin{figure}[h]
 \centering
 \includegraphics[width=\linewidth]{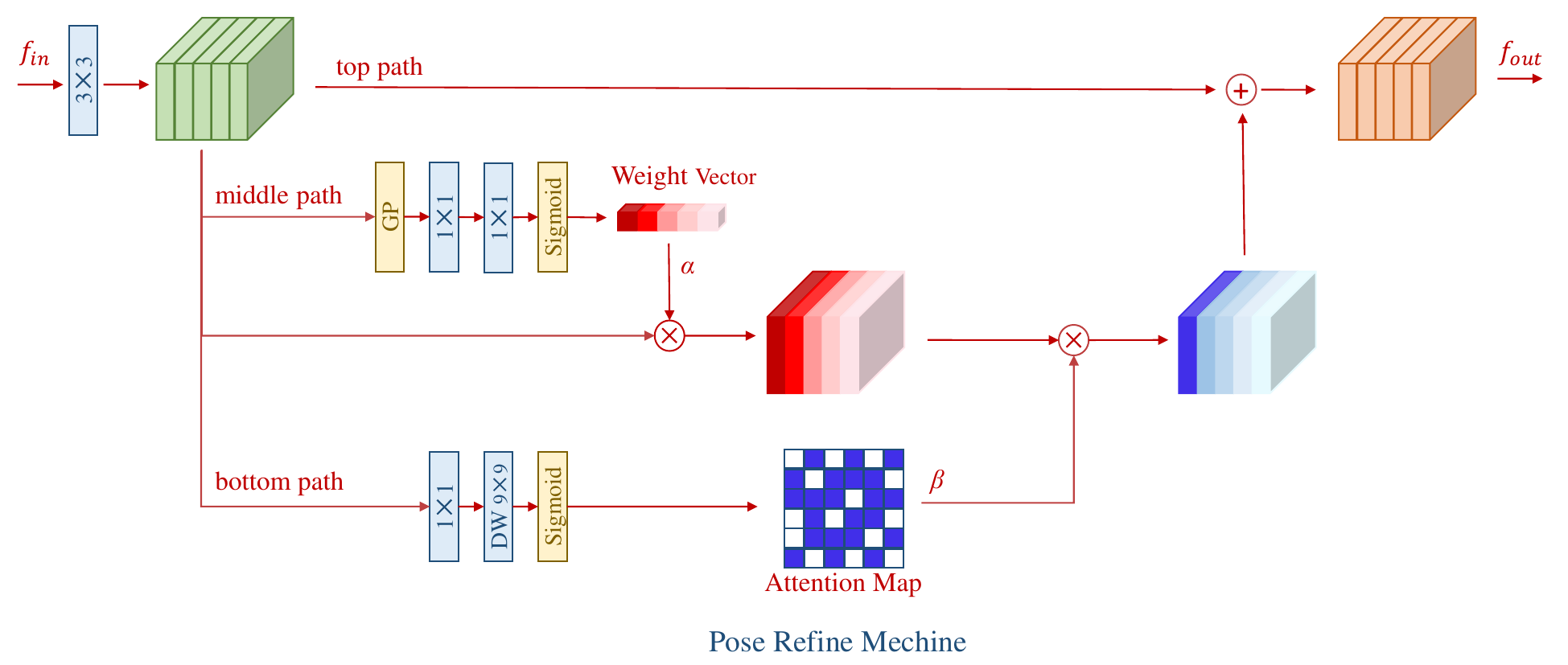} 
 \caption{Architecture of Pose Refine Machine (PRM). GP denotes global pooling. DW denotes depth-wise separable convolution. $\alpha$ denotes the weight vector. $\beta$ denots the attention map. The top path is an identity connection, the middle path is designed to reweight features in channel wise and the bottom path is proposed for spatial attention.} \vspace{-3mm}
 \label{fig:RB&FB} 
\end{figure}

\subsection{Pose Refine Machine} \label{sec:RM}
In the last module of multi-stage network (Figure~\figtabref{fig:pipeline}(a)), an attention mechanism - Pose Refine Machine (PRM) is used to reweight the output features, as shown in Figure~\figtabref{fig:RB&FB}. The first component of PRM is a conv3$\times$3, then the features are input into three paths. The top path is an identity connection. The middle one, motivated by SENet~\cite{senet}, passes through a global pooling, two conv1$\times$1 and a sigmoid activation to get a weight vector $\alpha$. The bottom path passes through a conv1$\times$1, a depth-wise separable conv9$\times$9 and a sigmoid activation to get an attention map $\beta$. Element-wise sum and multiplication are conducted among the three paths to get the output features.
Define the input features of PRM as $f_{in}$, the output features as $f_{out}$, the first conv3$\times$3 as $K(\cdot)$ and element-wise multiplication as $\odot$. Then  PRM can be formulated as Equation~\ref{eq_rm}.

\begin{equation}
  f_{out} = K(f_{in})\odot(1+\beta\odot\alpha)
  \label{eq_rm}
\end{equation}

As for the output of RSN, features after intra-level and inter-level aggregation are mixed together containing both low-level precise spatial information and high-level discriminant semantic information. Spatial information is good for keypoint localization while semantic information benefits keypoints classification. These features contribute differently to the final prediction. Therefore, to tackle this imbalance problem, PRM is designed to make a trade-off between local and global representations in output features of RSN. Compared to prior work of attention mechanism, we use powerful while computation-economical operations, e.g. conv3$\times$3, conv1$\times$1 and DW conv9$\times$9.  The top identity mapping in PRM is good for retaining local features which benefits precise keypoint localization. The middle path is designed to reweight the features in channel wise and the bottom path is proposed for spatial attention.


\vspace{-2mm}
\section{Experiments} \label{sec:experiment}
\vspace{-1mm}
\subsection{COCO Keypoints Detection}\label{sec:coco}
\vspace{-1mm}
\textbf{Datasets, Evaluation Metric, Human Detection.} COCO dataset~\cite{hrnet36} includes over 200K images and 250K person instances labeled with 17 joints. We use only COCO train2017 dataset for training (including about 57K images and 150K person instances). We evaluate our method on COCO minival dataset~(5K images) and the testing datasets including test-dev~(20K images) and test-challenge~(20K images). We use standard OKS-based AP score as the evaluation metric. We use MegDet and MegDet-v2 as human detecor on COCO val and test sets respectively.

\textbf{Training Details.} The network is trained on 8 V100 GPUs with mini-batch size 48 per GPU. There are 140k iterations per epoch and 200 epochs in total. Adam optimizer is adopted and the linear learning rate gradually decreases from 5e-4 to 0. The weight decay is 1e-5. Each image goes through a series of data augmentation operations including cropping, flipping, rotation and scaling. The range of rotation is $-45^{\circ} \sim +45^{\circ}$. The range of scaling is 0.7$\sim$1.35. The size of input image is 256$\times$192 or 384$\times$288.

\textbf{Testing Details.} We apply a post-Gaussian filter to the estimated heatmaps. Following the strategy of hourglass~\cite{hourglass}, we average the predicted heatmaps of original image with the results of corresponding flipped image. Then we implement a quarter offset from the highest response to the second highest one to get the locations of keypoints. The same with CPN~\cite{cpn}, the pose score is the multiplication of the average score of keypoints and the bounding box score.

\vspace{-3mm}
\begin{table}[h]
  \footnotesize
  \centering
  \caption{Results of ResNet, Res2Net, Baseline1/2, RSN, and RTMPose-X on val set.}
  \begin{tabular}{c|c|c|c|c}
    \hline\hline
    \quad ~~~~~model~~~~~~~~\quad &~~~\quad input size ~\quad\quad &~~~\quad AP ~\quad\quad &~~~~~~ $\Delta$ ~~~~~~&~~\quad GFLOPs~\quad\quad \\
    \hline
    ResNet-18 & 256$\times$192 & 70.7 & 0 & 2.3                   \\
    Res2Net-18 & 256$\times$192 & 71.3 & +0.6 & 2.2                   \\
    Baseline1-18 & 256$\times$192 & 72.9 & +2.1 & 2.5                   \\
    Baseline2-18 & 256$\times$192 & 72.1 & +1.4 & 2.5                   \\
\textbf{RSN-18} & \textbf{256$\times$192} & \textbf{73.6} &\textbf{+2.9} & \textbf{2.5}                   \\
    \hline
    ResNet-50 & 256$\times$192 & 72.2 & 0 & 4.6                   \\
    Res2Net-50 & 256$\times$192 & 72.8 & +0.6 & 4.5                   \\
    Baseline1-50 & 256$\times$192 & 73.7 & +1.5 & 6.4                   \\
    Baseline2-50 & 256$\times$192 & 72.7 & +0.5 & 6.4                   \\
\textbf{RSN-50} & \textbf{256$\times$192} & \textbf{74.7} &\textbf{+2.5}  & \textbf{6.4}                   \\
    \hline
    ResNet-101  & 256$\times$192 &73.2 & 0 &7.5 \\
    Res2Net-101  & 256$\times$192 &73.9 & +0.7 &7.5 \\
\textbf{RSN-101}   & \textbf{256$\times$192} &\textbf{75.8} &\textbf{+2.5}   &\textbf{11.5} \\
    \hline    
    4$\times${\rm ResNet-50} & 256$\times$192 &76.8 &0 &20.6 \\
    4$\times${\rm Res2Net-50} & 256$\times$192 &77.0 &+0.2 &20.1 \\
    4$\times${\rm \textbf{RSN-50}} & \textbf{256$\times$192} &\textbf{78.6} &\textbf{+1.8} &\textbf{27.5} \\
    \hline
    4$\times${\rm ResNet-50} & 384$\times$288 &77.5 &0 &46.4 \\
    4$\times${\rm Res2Net-50} & 384$\times$288 &77.6 &+0.1 &45.2 \\
    4$\times${\rm \textbf{RSN-50}} & \textbf{384$\times$288} &\textbf{79.2} &\textbf{+1.7} &\textbf{61.9} \\
    \hline
    RTMPose-X & 384$\times$288 & 78.8 & 0 & 17.2 \\
    RTMPose-X + Res2Net & 384$\times$288 & 78.7 & $-$0.1 & 17.1 \\
    \textbf{RTMPose-X + RSN} & \bf 384$\times$288 & \textbf{79.1} & \textbf{+0.3} &\bf 29.6 \\
    \hline\hline
  \end{tabular}
  \vspace{-6mm}
  \label{tb:qvxian}
\end{table}
\begin{figure}[h]

 \centering
 \includegraphics[width=1.0\textwidth]{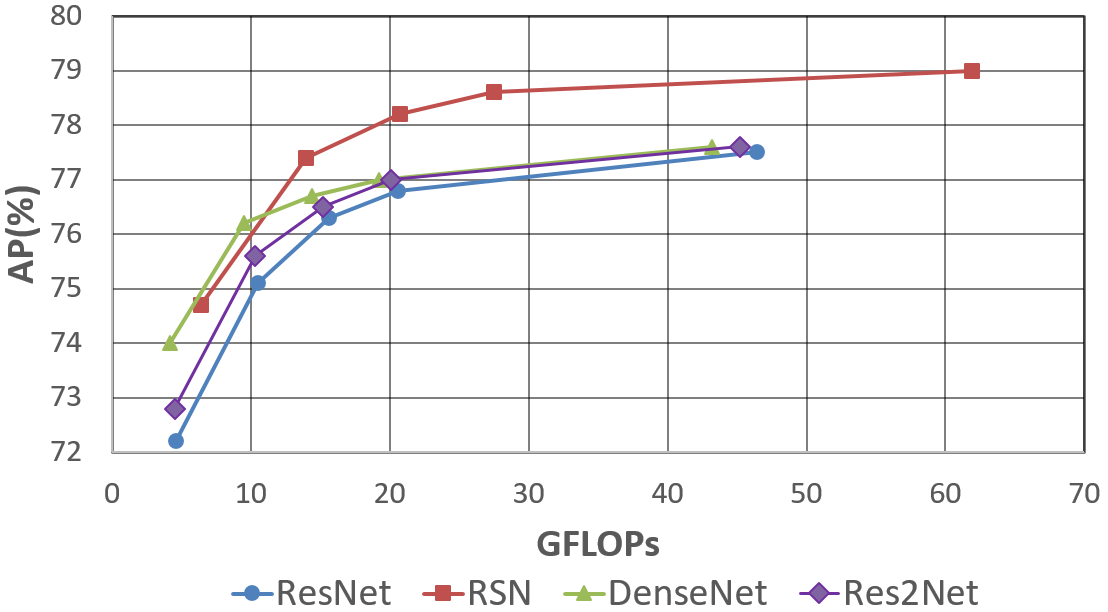} 
 \caption{Illustrating how the performances of ResNet, Res2Net, DenseNet and RSN are affected by GFLOPs. The results are reported on COCO minival dataset.} 
 \label{fig:qvxian2} 
\end{figure}

\subsubsection{Ablation Study of RSN Improvement.}\label{sec: ablation-rsn}
In Section~\ref{sec:fusion}, we analyze the differences between RSN and current methods. In this part, we validate the effectiveness of intra-level feature fusion method in RSN. Since it is known dividing the network into branches, e.g., Inception and ResNetXt~\cite{resnext}, can improve the recognition performance,  we add two baselines into comparison. Baseline 1: remove the intra-level fusion (i.e., the vertical arrows) from Figure~\figtabref{fig:pipeline}(c).  This baseline can reveal whether the proposed intra-level fusion is important. Baseline 2: replace f1-f4 in Baseline 1 with  4 f3's respectively, which is more like the conventional branching strategy. We keep the same GFLOPs of Baseline1, Baseline2 with RSN by adapting channels. Ablation experiments are implemented on ResNet, Res2Net, Bseline1, Baseline2, and RSN based networks. PRM is left out for more strong comparison.  The results on COCO val are reported in Table~\figtabref{tb:qvxian}.

As Table~\figtabref{tb:qvxian} shows, RSN boosts the performance by relatively larger extent with acceptable computation cost addition, while Res2Net can only obtain limited gain. For instance, RSN-18 is 2.9 points AP higher than ResNet-18 adding only 0.2 GFLOPs and 2.3 points AP higher than Res2Net-18 adding only 0.3 GFLOPs. However, Res2Net-18 obtains only 0.6 AP gain than ResNet-18. On RTMPose-X, Res2Net degrades the baseline by 0.1 AP while RSN improves the performance by 0.3 AP. 
RSN always works much better than ResNet and Res2Net with comparable GFLOPs. In addition, it is worth noting that when model complexity is relatively low, RSN still has a remarkable performance, which indicates that RSN is more compact and efficient. For instance, compared with ResNet-101 and Res2Net-101, RSN-18 has a similar AP, however, with only a third of computation cost. On the other hand, RSN achieves higher AP than Baseline1 and Baseline2 with the same GFLOPs, e.g., RSN-50 is 1 AP higher than Baseline1-50 and 2 AP higher than Baseline2-50. This observation strongly demonstrate the superiority of the intra-level feature fusion mode of RSN.

\subsubsection{Ablation Study of RSN Efficiency.}\label{sec:high-efficiency}
The dense connection principle of RSN comes form DenseNet. However, it is not efficient for DenseNet when too many concatenating operations are implemented. To circumvent the network capacity explosion, RSN uses element-wise sum to connect adjacent branches. To validate the efficiency of our approach, we respectively adopt ResNet, Res2Net, DenseNet and RSN as the backbone in the same multi-stage architecture as shown in Figure~\figtabref{fig:pipeline}(a) to compare the performance. PRM is left out for fair comparison. The results are shown in Figure~\figtabref{fig:qvxian2}. For relatively small models, RSN and DenseNet based networks can both achieve good results, while Res2Net only gets a minor improvement than ResNet. However, as the model capacity increases, the improvements of DenseNet and Res2Net based network decrease dramatically. Both of them can only get a inferior result close to ResNet when the model size becomes large, while RSN can keep its superiority to the end.

DenseNet has a high AP score at a low complexity owing to the deep connections and frequent feature aggregations inside the same level by continuous concatenating operations. This makes the low-level features sufficiently supervised resulting in satisfactory delicate spatial texture information, which benefits the keypoint localization. However, as the computation cost raises, the concatenating operations of DenseNet become redundant. It combines quite a large range of less utilized information. As for Res2Net, narrower range of receptive fields and lack of efficient intra-level feature fusion to extract delicate local representations make it much inferior than RSN.

\begin{figure}[h]

 \centering
 \includegraphics[width=1.0\textwidth]{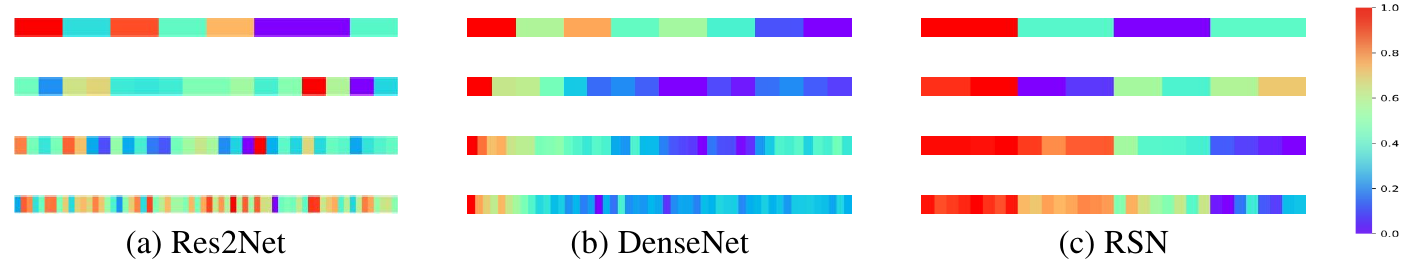} 
 \caption{The average absolute filter weights of the last conv$1\times1$ layers of each level in trained Res2Net-50 (a), DenseNet-169(b) and RSN-50(c). Larger weights means higher utilization. The weights of Res2Net are smaller than those of RSN. Most weights in DenseNet have values close to zero. While RSN can utilize most channels better.} 
 \label{fig:weights} 
\end{figure}

\begin{figure}[h]

 \centering
 \includegraphics[width=1.0\textwidth]{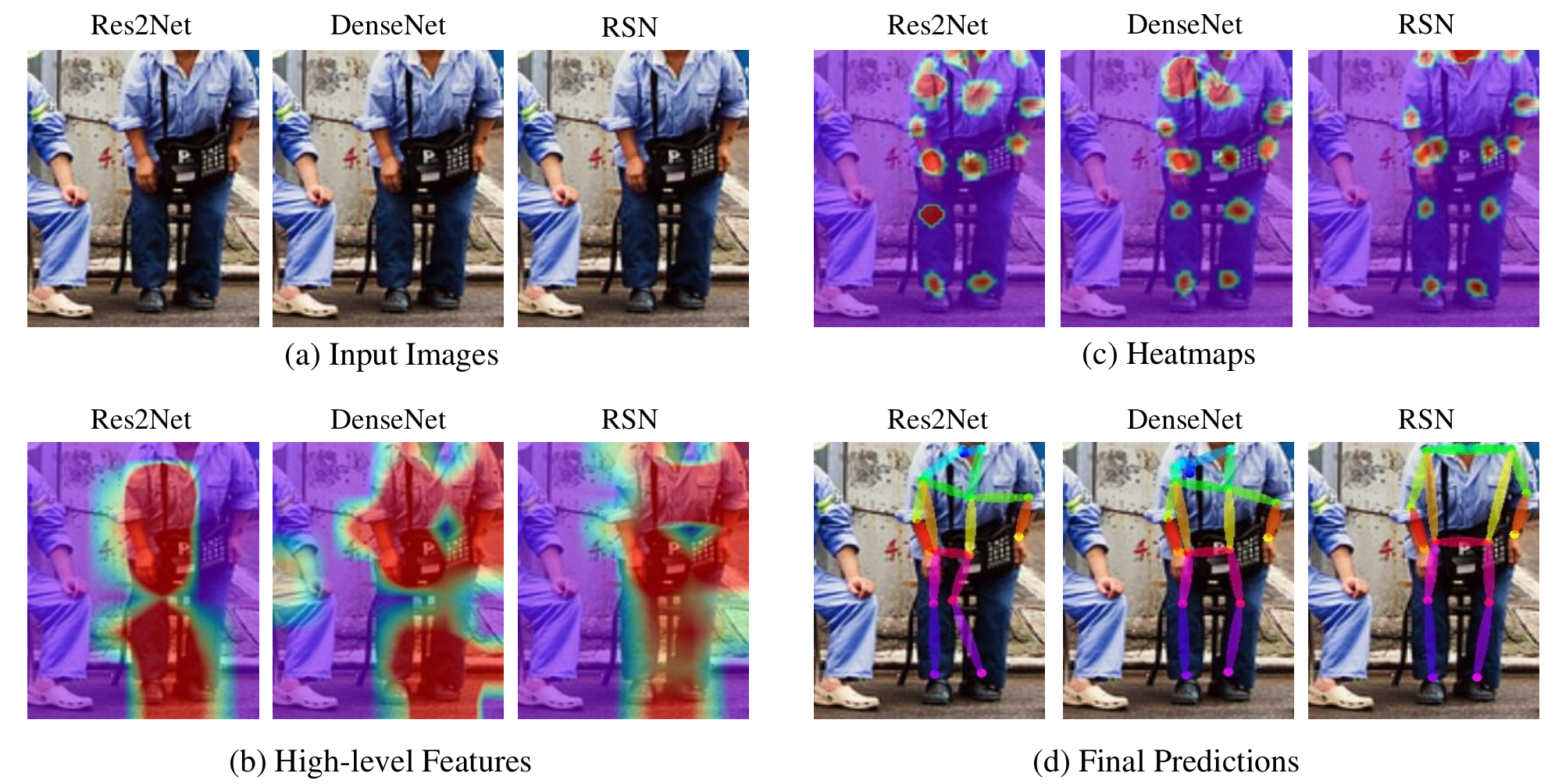} 
 \caption{Visual analysis of Res2Net-50, DenseNet-169 and RSN-50. (a) Input images, (b) High-level feature maps (level 4), (c) Low-level heatmaps (level 1), (d) Final predictions.} 
 \label{fig:feature} 
\end{figure}

In order to embody the differences of Res2Net, DenseNet and RSN more essentially, we show the average absolute filter weights of the last conv1$\times$1 layers of each level in trained Res2Net-50, DenseNet-169 and RSN-50 in Figure~\figtabref{fig:weights}. The highly used weights become less from level 1 to level 4 in DenseNet. The overall useful weights of DenseNet are less than those of RSN, which can be deduced from Figure~\figtabref{fig:weights} (b) and (c) that the red area in each level of DenseNet is much smaller than that of RSN. According to the analysis in Section~\ref{sec:recept-filed}, RSN can enhance the efficient fusion of intra-level features with dense element-wise sum connections. There are not accumulative concatenating operations like DenseNet. Thus, RSN is less occupied by the redundant features with low utilization. On the other hand, compared with Res2Net, the more densely connected architecture and wider range of receptive fields make the intra-level feature fusion of RSN more effective, that is why the red area of RSN is much larger than that of Res2Net and the weights of RSN are more fully used, just as shown in Figure~\figtabref{fig:weights} (a) and (c). As a result, the RSN model can keep its high efficiency and considerable improvement from the beginning to the end, just as shown in Figure~\figtabref{fig:qvxian2}.



Additionally, to highlight the advantages of RSN more intuitively, we conduct visual analysis of Res2Net-50, DenseNet-169 and RSN-50, as shown in Figure~\figtabref{fig:feature}. In Figure~\figtabref{fig:feature}(b), the high-level response to human body of Res2Net and DenseNet either covers incomplete body area or too large area of background. Only RSN has a relatively complete and appropriate response area to the human body. As a result, in final prediction, Res2Net is easily misled by the background information, DenseNet ignores some keypoints such as shoulders, while RSN can locate the keypoints better and reduce the interference of background information. As Figure~\figtabref{fig:feature}(c) shows, the heatmaps of RSN are much clearer and the locations of the responses are much more accurate.

\subsubsection{Ablation Study of RSN Architecture.}\label{sec:branch}
When designing RSN, we firstly deploy the dense connection principle of DenseNet. Then, for a break-down ablation, we set different branch number as variants to discuss the designing of RSN and explore a best trade-off between branch representing capacity and the degree of intra-level fusion. experiments are done on RSN-18 and RSN-50. We increase branch numbers from $2$ to $6$ while keeping the model capacity unchanged by adapting channels. As Table \figtabref{tb:branch} shows, the performance firstly becomes better and attains its peak when there are $4$ branches. However, when the branch number continues growing, the results get worse. Thus, $4$ is the best choice.
\begin{table}[h]

  \centering
  \caption{Illustrating how the performance of RSN affected by the branch number.}
  \begin{tabular}{c|c|c|c|c|c|c}
    \hline\hline
    ~~backbone ~~& ~~input size ~~&~ 2-branch ~&~ 3-branch ~& ~4-branch ~& ~5-branch~ &~ 6-branch~ \\
    \hline
    RSN-18 &256$\times$192 &73.1 & 73.0  &\textbf{73.6} &73.2 &72.9        \\
    RSN-50 &256$\times$192 &73.9 & 74.2  &\textbf{74.7} &74.3 &74.0        \\
    \hline\hline
  \end{tabular}
  
  \label{tb:branch}
\end{table}

\subsubsection{Ablation Study of Pose Refine Machine.} \label{sec:rb_ffb}
In Section~\ref{sec:RM}, we have analyzed the architecture of PRM and the differences between PRM and prior attention mechanism. To validate the improvement of PRM, we perform ablation experiments on both single-stage and multi-stage network architecture. Additionally, We validate the impact of SENet and CBAM by replacing PRM. The results are shown in Table~\figtabref{tb:rb_ffb}. SE-block and CBAM decrease the performance of human pose estimation, which implies vanilla attention mechanisms are not suitable for rebalancing output features. In contrast, when the model capacity is small, PRM has a considerable improvement. As for relatively high AP baseline, PRM still obtains 0.4 AP gain. These observations demonstrate the robustness of PRM.  

\subsubsection{Results on COCO test-dev and test-challenge.}\label{sec:ep_final}
We validate our approach on COCO test-dev and test-challenge sets. The results are shown in Table~\figtabref{tb:test-dev} and Table~\figtabref{tb:test-challenge}. For fair comparison, we pay attention to the performances of  single models with comparable GFLOPs, without using extra training data. In Table~\figtabref{tb:test-dev}, our method outperforms HRNet by 2.5 AP (78.0 v.s. 75.5), and outperforms SimpleBaseline by 4.3 AP on COCO test-dev dataset. Additionally, as Table~\figtabref{tb:test-challenge} shows, our approach outperforms MSPN (winner of COCO kps Challenge 2018) by 0.7 AP on test-challenge set. Note that we don't even use pretrained model.

\begin{table}[h]
  \footnotesize

  \centering
  \caption{Ablation experiments of Pose Refine Machine on COCO minival dataset.}
  \begin{tabular}{c|c|c|c|c|c}
    \hline\hline
    ~~~~~Backbone ~~~~~&~~ Attention ~~&~~ input size ~~\quad&~~\quad AP ~\quad\quad &~~~~$\Delta$~~~~ &~~ GFLOPs ~\quad \\
    \hline
    ResNet-18 &None &256$\times$192 & 70.7 &0 &2.3 \\
    ResNet-18 &SE-block &256$\times$192 & 70.5 &-0.2 &2.3 \\
    ResNet-18 &CBAM &256$\times$192 & 69.9 &-0.8 &2.3 \\
    \textbf{ResNet-18} &PRM &256$\times$192  & \textbf{72.2} &\textbf{+1.5} & \textbf{4.1}   \\
    \hline
    ResNet-50 &None &256$\times$192 & 72.2 &0 &4.6 \\
    ResNet-50 &SE-block &256$\times$192 & 72.1 &-0.1 &4.6 \\
    ResNet-50 &CBAM &256$\times$192 & 71.1 &-1.1 &4.6 \\
    \textbf{ResNet-50} &PRM &256$\times$192 & \textbf{73.4} &\textbf{+1.2} & \textbf{6.4}   \\
    \hline
    4$\times$ResNet-50 &None &256$\times$192 & 76.8 &0 &20.6 \\
    4$\times$ResNet-50 &SE-block &256$\times$192 & 76.6 &-0.2 &20.6 \\
    4$\times$ResNet-50 &CBAM &256$\times$192 & 76.1 &-0.7 &20.6 \\
    \textbf{4$\times$ResNet-50} &PRM &256$\times$192 & \textbf{77.2} &\textbf{+0.4} & \textbf{22.4}   \\ 
    \hline     
    4$\times$RSN-50 &None &256$\times$192 & 78.6 &0  &27.5                   \\
    4$\times$RSN-50 &SE-block &256$\times$192 & 78.6 &0  &27.5                   \\
    4$\times$RSN-50 &CBAM &256$\times$192 & 78.0 &-0.6  &27.5                   \\
    \textbf{4$\times$RSN-50} &PRM &256$\times$192 & \textbf{79.0} &\textbf{+0.4} & \textbf{29.3}   \\   
    \hline\hline
  \end{tabular}
  
  \label{tb:rb_ffb}
\end{table}

\begin{table*}[h]
  \tiny

  \centering
  \caption{Results on COCO test-dev dataset. "*" means using ensembled models. Pretrain = pretrain the backbone on the ImageNet classification task.}
   \resizebox{\textwidth}{23mm}{
  \begin{tabular}{c|c|c|c|c|c|c|cccccc}
    \hline
    \hline
    Method &Extra data &Pretrain & Backbone & Input Size &Params & GFLOPs & AP & AP.5 & AP.75 &AP(M) &AP(L) &AR \\ 
    \hline
    CMUpose~\cite{cmu-pose} &$\times$          &- &- &- &- &- & 61.8 &84.9 &67.5 &57.1 &68.2 &66.5                  \\
    
    G-MRI~\cite{GMRI} &$\times$ &- & ResNet-101       &353$\times$257 &42.6M  &57.0 &64.9 &85.5 &71.3 &62.3 &70.0 &69.7 \\  
    
G-RMI~\cite{GMRI} &$\surd$ &- &ResNet-101     &353$\times$257 &42.6M &57.0 &68.5 &87.1 &75.5 &65.8 &73.3 &73.3 \\

CPN~\cite{cpn} &$\times$ &$\surd$ &ResNet-Inception          &384$\times$288 &58.8M  &29.2 &72.1 &91.4 &80.0 &68.7 &77.2 &78.5 \\

CPN$^{*}$~\cite{cpn} &$\times$ &$\surd$  &ResNet-Inception    &384$\times$288 &- &- &73.0 &91.7 &80.9 &69.5 &78.1 &79.0 \\

SimpleBase~\cite{simplebase} &$\times$ &$\surd$ &ResNet-152       &384$\times$288 &68.6M &35.6 &73.7 &91.9 &81.1 &70.3 &80.0 &79.0 \\

HRNet-W32~\cite{hrnet} &$\times$ &$\surd$ &HRNet-W32          &384$\times$288 &28.5M &16.0 &74.9 &92.5 &82.8 &71.3 &80.9 &80.1 \\

HRNet-W48~\cite{hrnet} &$\times$ &$\surd$ &HRNet-W48          &384$\times$288 &63.6M &32.9 &75.5 &92.5 &83.3 &71.9 &81.5 &80.5 \\

SimpleBase$^{*}$\cite{simplebase} &$\surd$ &$\surd$ &ResNet-152   &384$\times$288  &- &- &76.5 &92.4 &84.0 &73.0 &82.7 &81.5 \\

HRNet-W48~\cite{hrnet} &$\surd$&$\surd$ &HRNet-W48    &384$\times$288 &63.6M &32.9 &77.0 &92.7 &84.5 &73.4 &\textbf{83.1} &82.0 \\
MSPN~\cite{mspn} &$\surd$&$\times$ &4$\times$ResNet-50   &384$\times$288 &71.9M &58.7 &77.1 &93.8 &84.6 &73.4 &82.3 &82.3  \\
    \hline
Ours(RSN) &$\times$&$\times$ &RSN-18  &256$\times$192 &12.5M &2.5 &71.6 &92.6 &80.3 &68.8 &75.8 &77.7 \\
Ours(RSN) &$\times$&$\times$ &RSN-50  &256$\times$192 &25.7M &6.4 &72.5 &93.0 &81.3 &69.9 &76.5 &78.8 \\
Ours(RSN) &$\times$&$\times$ &2$\times$RSN-50  &256$\times$192 &54.0M &13.9 &75.5 &93.6 &84.0 &73.0 &79.6 &81.3 \\
\textbf{Ours(RSN)} &$\times$&$\times$ &\textbf{4$\times$RSN-50}  &256$\times$192 &111.8M &\textbf{29.3} &\textbf{78.0} &\textbf{94.2} &\textbf{86.5} &\textbf{75.3} &82.2 &\textbf{83.4} \\
\textbf{Ours(RSN)}&$\times$&$\times$ &\textbf{4$\times$RSN-50} &384$\times$288 &111.8M &\textbf{65.9} &\textbf{78.6} &\textbf{94.3} &\textbf{86.6} &\textbf{75.5} &\textbf{83.3} &\textbf{83.8} \\
\textbf{Ours(RSN$^{*}$)} &$\times$&$\times$ &\textbf{4$\times$RSN-50} &- &- &-&\textbf{79.2} &\textbf{94.4} &\textbf{87.1} &\textbf{76.1} &\textbf{83.8} &\textbf{84.1} \\
\hline
\hline
  \end{tabular}
  }
  
  \label{tb:test-dev}
\end{table*}

\subsubsection{Inference Speed.}\label{sec:inference_speed}
Current methods of human pose estimation mainly focus on promoting the performance while deploying resource-intensive networks with large depth and width. This leads to inefficient inference. Interestingly, we observe RSN can make a better trade-off between accuracy and inference speed than prior work.  For fair comparison, we train RSN and HRNet under the same settings in Section~\ref{sec:experiment}, with 256$\times$192 input size. Both use MegDet as   human detector when testing. We use pps to measure inference speed, i.e., Persons inferred Per Second. On the same GPU (RTX 2080ti), results of COCO val are reported, HRNet-w16 with 1.9 G and 7.2 M achieves 71.9 AP and 31.8 pps, RSN-18 with 2.5G and 12.5 M achieves 73.6 AP and
64.9 pps, HRNet-w32 with 7.1 G and 28.5M achieves 74.6 AP and 26.5 pps, RSN-50 with 6.4 G and 25.7 M achieves 74.7 AP and 42.6 pps. HRNet-w48 with 14.6G and 63.6M achieves 75.5 AP and 24.7 pps, 2$\times$RSN-50 with 13.9 G and 54.0 M achieves 77.4 AP and 20.2 pps. In addition, the inference speed on CPU (Intel(R) Xeon(R) Gold6013@2.1GHZ) also shows, RSNs with higher performances are faster than HRNet by all sizes. These results suggest that RSN is more accurate, compact and efficient.

\subsubsection{Effect of Human Detection.}\label{sec:human_det}
We use MegDet as human detector in ablation study, which achieves 49.4 AP on COCO val. For test sets, we use MegDet-v2, which has 59.8 AP on COCO val. As human detection has an influence on the final performance of top-down  approach, We perform ablations to investigate the impact of human detector on COCO test-dev.
4$\times$RSN-50 at input size of 256$\times$192 achieves 77.3 AP using MegDet, and  78.0 using MegDet-v2. 4$\times$RSN-50 at input size of 384$\times$288 achieves 77.9 using MegDet, and 78.6 using MegDet-v2.

\begin{table*}[h]
  \tiny
  \caption{Results on COCO test-challenge dataset. "*" means using ensembled models.}
  \centering
  \resizebox{\textwidth}{11mm}{
  \begin{tabular}{c|c|c|c|c|c|c|cccccc}
    \hline
    \hline
    Method & Extra data &Pretrain & Backbone & Input Size &Params &GFLOPs  & AP & AP.5 & AP.75 &AP(M) &AP(L) &AR \\ 
    \hline
    G-RMI~\cite{GMRI} &$\surd$ &- &ResNet-101                 &353$\times$257 &42.6M &57.0 &69.1 &85.9 &75.2 &66.0 &74.5 &75.1  \\
    CPN$^{*}$~\cite{cpn} &$\times$ &$\surd$ &ResNet-Inception          &384$\times$288 &- &-  &72.1 &90.5 &78.9 &67.9 &78.1 &78.7  \\
    Sea Monsters$^{*}$ &$\surd$ &- &-  &- &-   &- &74.1 &90.6 &80.4 &68.5 &82.1 &79.5  \\
    SimpleBase$^{*}$\cite{simplebase} &$\surd$ &$\surd$ &ResNet-152   &384$\times$288 &- &-  &74.5 &90.9 &80.8 &69.5 &82.9 &80.5 \\
    MSPN$^{*}$~\cite{mspn} &$\surd$ &$\times$ &4$\times$ResNet-50   &384$\times$288 &- &- &76.4 &92.9 &82.6 &71.4 &83.2 &82.2  \\
    \hline
\textbf{Ours(RSN$^{*}$)} &$\times$&$\times$ &\textbf{4$\times$RSN-50}  &- &- &-  &\textbf{77.1} &\textbf{93.3} &\textbf{83.6} &\textbf{72.2} &\textbf{83.6} &\textbf{82.6} \\
  \hline
  \hline
  \end{tabular}
  }
  \label{tb:test-challenge}
\end{table*}

\subsection{MPII Human Pose Estimation}\label{sec:mpii}
We validate RSN on MPII test set, a  single-person pose estimation benchmark. As shown in Table~\figtabref{tb:mpii}, RSN boosts the SOTA performance by 0.7 in PCKh@0.5, which demonstrates the superiority and generalization ability of our method.

\begin{table}[tp]
  \tiny

  \centering
  \caption{PCKh@0.5 results on MPII test dataset.}
  \resizebox{\textwidth}{14mm}{
  \begin{tabular}{c|ccccccc|c}
    \hline
    \hline
    ~~~~~~~~Method~~~~~~~~ & ~~Hea~~ & ~~Sho~~ & ~~Elb~~ &~~Wri~~ &~~Hip~~ &~~Kne~~ &~~Ank~~ &~~~Mean~~~ \\
    \hline
    Chen et al.\cite{Chen_2017_ICCV} &98.1	&96.5	&92.5	&88.5	&90.2	&89.6	&86.0	&91.9 \\
    Yang et al.\cite{lpf} &98.5	&96.7	&92.5	&88.7	&91.1	&88.6	&86.0	&92.0 \\
    Ke et al.~\cite{Ke_2018_ECCV}  &98.5	&96.8	&92.7	&88.4	&90.6	&89.3	&86.3	&92.1 \\
    Tang et al.~\cite{hrnet62} &98.4	&96.9	&92.6	&88.7	&91.8	&89.4	&86.2	&92.3 \\
    Sun et al.~\cite{hrnet} &\textbf{98.6} &96.9 &92.8 &89.0 &91.5 &89.0 &85.7 &92.3 \\
    \hline
    
    \textbf{ours(4$\times$RSN-50)}  &98.5   &\textbf{97.3}   &\textbf{93.9}    &\textbf{89.9}  &\textbf{92.0}  &\textbf{90.6} &\textbf{86.8}   &\textbf{93.0} \\
    \hline
    \hline
  \end{tabular}
  }
  
  \label{tb:mpii}
\end{table} 

\section{Conclusion}
\begin{figure*}[h]
 \centering

 \includegraphics[width=0.95\textwidth]{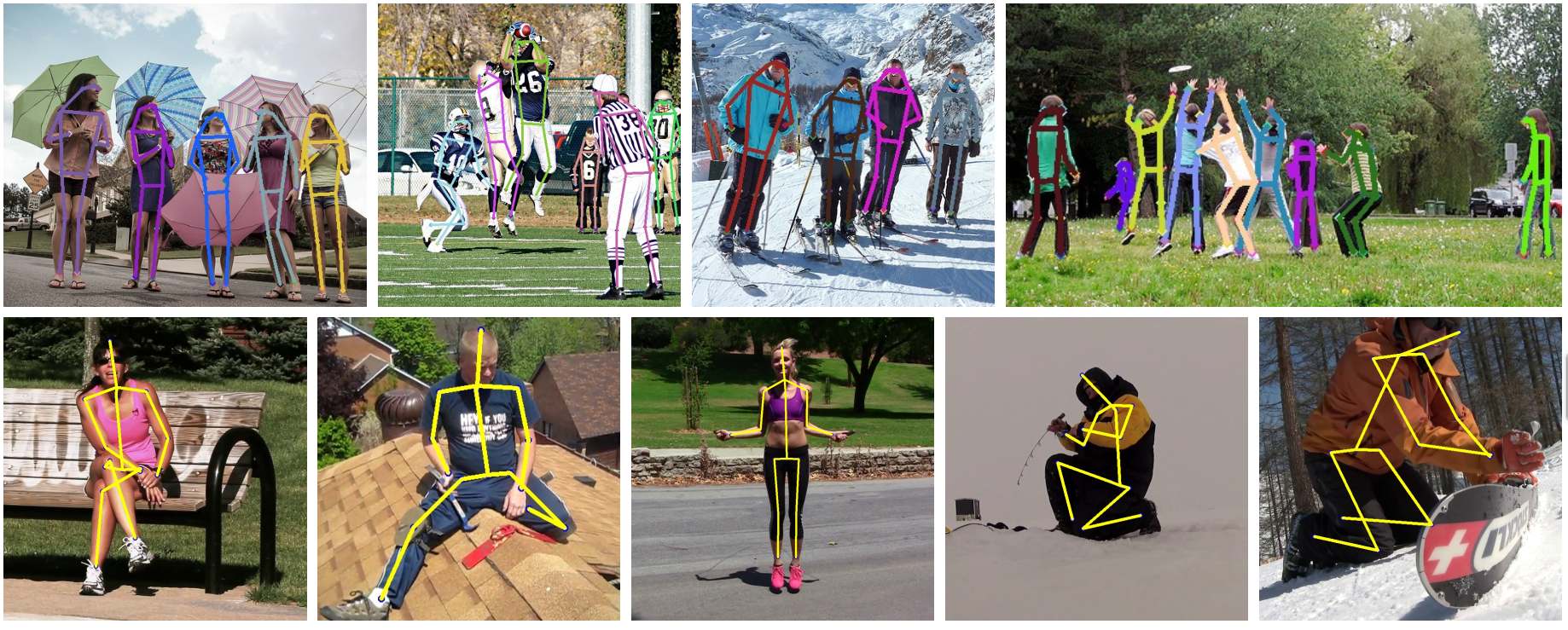} 
 \caption{Prediction results on COCO (top line) and MPII (bottom line) val sets. } 
 \label{fig:results_coco} 
\end{figure*}

In this paper, we propose a novel method, Residual Steps Network, which aims to learn delicate local representations by efficient intra-level feature fusion. To make a better trade-off between local and global representations in output features, we design Pose Refine Machine. Our method yields the best results on two benchmarks, COCO and MPII. Some prediction results are visualized in Fig~\figtabref{fig:results_coco}.

\section{Acknowledgement}
This work was supported in part by the National Key Research and Development Program of China under Grant 2017YFA0700800.

\clearpage
%
%
\bibliographystyle{splncs04}
\bibliography{egbib}
\end{document}